\documentclass{article}
\usepackage{microtype}
\usepackage{graphicx}
\usepackage{subfigure}
\usepackage{booktabs} 

\usepackage{hyperref}

\usepackage[accepted]{icml2023}

\usepackage{amsmath}
\usepackage{amssymb}
\usepackage{mathtools}
\usepackage{amsthm}

\usepackage[capitalize,noabbrev]{cleveref}
\usepackage{makecell}
\usepackage[mathscr]{eucal}
\usepackage{amsfonts,amssymb} 
\usepackage{booktabs}
\usepackage{stfloats}
\usepackage{float}
\usepackage{amsmath}
\usepackage{multirow}
\usepackage{amssymb}
\usepackage{pdfpages}
\usepackage{subfigure}
\usepackage{tabularx}
\usepackage{ragged2e}
\theoremstyle{plain}

\theoremstyle{definition}

\theoremstyle{remark}

\usepackage[textsize=tiny]{todonotes}

\icmltitlerunning{MVKT-ECG: Efficient Single-lead ECG Classification on Multi-Label Arrhythmia by Multi-View Knowledge Transferring}

\begin{document}

\twocolumn[
\icmltitle{MVKT-ECG: Efficient Single-lead ECG Classification on Multi-Label \\
Arrhythmia by Multi-View Knowledge Transferring}



\icmlsetsymbol{equal}{*}

\begin{icmlauthorlist}
\icmlauthor{Yuzhen Qin}{Tsinghua University}
\icmlauthor{Li Sun}{Tsinghua University}
\icmlauthor{Hui Chen}{Tsinghua University}
\icmlauthor{Wei-qiang Zhang}{Tsinghua University}
\icmlauthor{Wenming Yang}{Tsinghua University}
\icmlauthor{Jintao Tao}{Tsinghua University}
\icmlauthor{Guijin Wang}{Tsinghua University}
\end{icmlauthorlist}

\icmlkeywords{Machine Learning, ICML}

\vskip 0.3in
]



\printAffiliationsAndNotice{\icmlEqualContribution} 

\begin{abstract}
The widespread emergence of smart devices for ECG has sparked demand for intelligent single-lead ECG-based diagnostic systems. However, it is challenging to develop a single-lead-based ECG interpretation model for multiple diseases diagnosis due to the lack of some key disease information. In this work, we propose inter-lead \textbf{M}ulti-\textbf{V}iew \textbf{K}nowledge \textbf{T}ransferring of ECG (MVKT-ECG) to boost single-lead ECG's ability for multi-label disease diagnosis. This training strategy can transfer superior disease knowledge from multiple different views of ECG (e.g. 12-lead ECG) to single-lead-based ECG interpretation model to mine details in single-lead ECG signals that are easily overlooked by neural networks. MVKT-ECG allows this lead variety as a supervision signal within a teacher-student paradigm, where the teacher observes multi-lead ECG educates a student who observes only single-lead ECG. Since the mutual disease information between the single-lead ECG and muli-lead ECG plays a key role in knowledge transferring, we present a new disease-aware \textit{Contrastive Lead-information Transferring}(CLT) to improve the mutual disease information between the single-lead ECG and muli-lead ECG. Moreover, We modify traditional Knowledge Distillation to \textit{multi-label disease Knowledge Distillation} (MKD) to make it applicable for multi-label disease diagnosis. The comprehensive experiments verify that MVKT-ECG has an excellent performance in improving the diagnostic effect of single-lead ECG.
\end{abstract}

\section{Introduction}
\begin{figure*}[t]
	\centering
	\subfigure[12-lead ECG from 12 different viewpoints]{
		\begin{minipage}[t]{0.73\textwidth}
			\centering
			\includegraphics[width=0.99\linewidth]{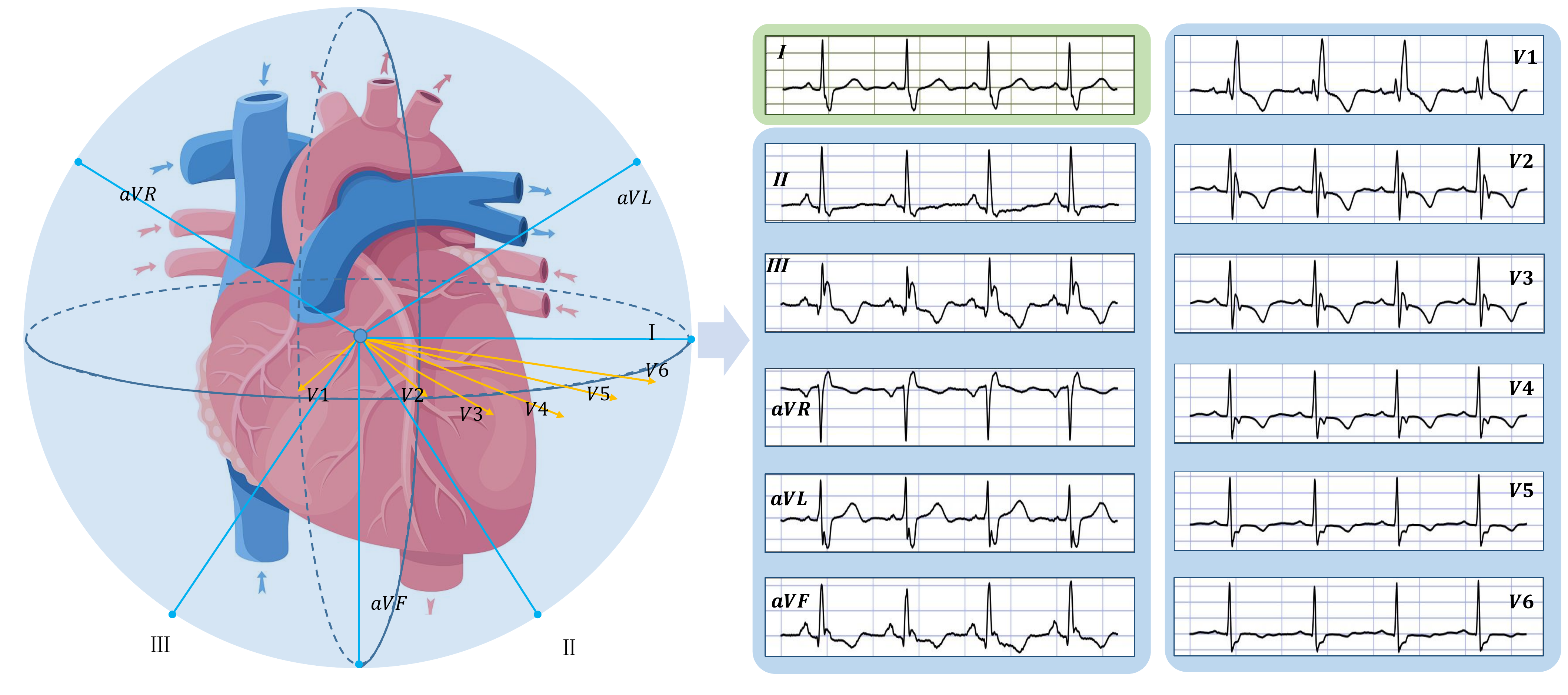}\\
            \label{fig1: sub_figure1}
		\end{minipage}%
	}%
	\subfigure[A ECG waveform with arrhythmia STD (ST-segment depression)]{
		\begin{minipage}[t]{0.27\textwidth}
			\centering
			\includegraphics[width=0.99\linewidth]{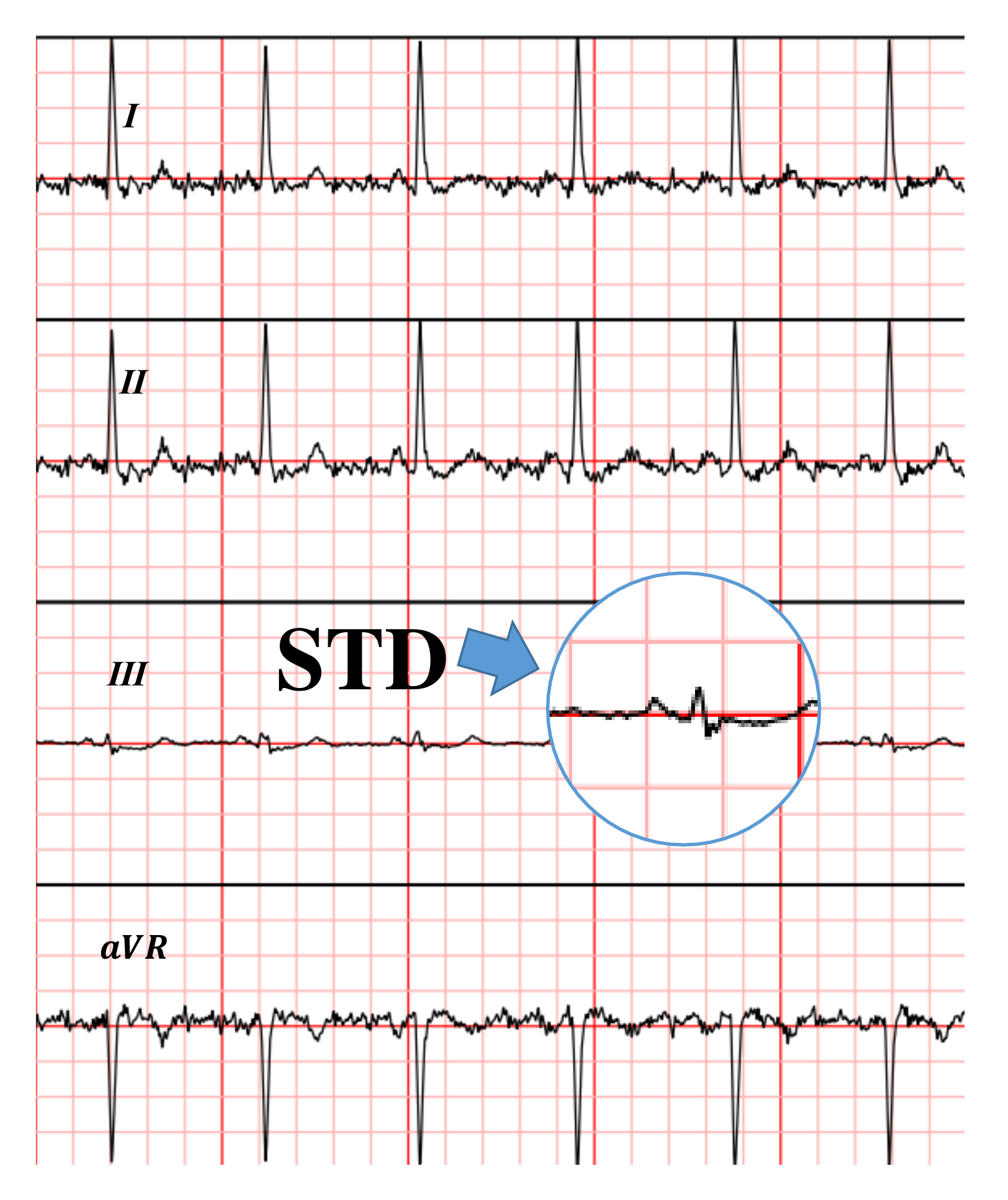}\\
            \label{fig1: sub_figure2}
		\end{minipage}%
	}%
	\centering
	\caption{(a): A standard 12-lead ECG signal by depicting heartbeat from 12 different views. Mul-view ECG can record the heart's activity more comprehensively, as it includes rich disease information from various perspective. (b): Lead III of this ECG signal shows the disease STD, while other leads are all normal (only four leads are shown), so judgment based on lead I or other lead (except lead III) alone is likely to lead to missed detection.}
    \label{fig1}
	\vspace{-0.1cm}
	\label{fig1}
\end{figure*}

Electrocardiogram (ECG) is a commonly used, non-invasive, and convenient diagnostic method for detecting Cardiac arrhythmia and other cardiovascular conditions \cite{holst1999confident}. It can be divided into two categories: Multi-lead (12-lead system standard in clinical) ECG and single-lead ECG. Multi-lead ECG is acquired by several electrodes arranged on the patient's chest wall and limbs, which can be seen as visualizing the heartbeat signals in multiple different views (Figure \ref{fig1: sub_figure1})  \cite{https://doi.org/10.48550/arxiv.2105.06293, pmlr-v162-chen22n}. Under the strong ability of deep learning, many multi-lead-based deep ECG interpretation models are widely used in practice. Previous models focused on extracting discriminative features \cite{7202837,ribeiro2020automatic,WANG2019523}. Subsequently, the residual connection is widely used in deeper convolutional network structures to increase the expressive power of the model\cite{ribeiro2020automatic}. To exploit the properties of different leads, a series of deep neural networks are proposed to establish an attention mechanism between leads \cite{10.1007/978-3-030-33327-0_10,bios11110453}. More recently, faced with the shortage of high-quality ECG datasets, many self-supervised approaches focus on mining effective information from unlabeled data \cite{kiyasseh2021clocs, 9543620, 9161416}. However, this kind of high-quality multi-view ECG data is provided by an environment with strong observation equipment (e.g. hospital) \cite{hong2018rdpd}.

Single-lead electrocardiograms have attracted much attention recently due to the emerging smart ECG devices, such as Apple Watch and Alivecore, which bring great convenience for people. In recent years, many single-lead ECG signals have been produced with the popularization of intelligent ECG devices, raising high expectations for robust intelligent multi-label disease single-lead ECG diagnostic systems. However, the poor observation equipment always provides low-quality and single-view ECG signals \cite{hong2018rdpd}. Due to the lack of views, research on the intelligent diagnosis of single-lead ECG still focuses on a few diseases\cite{RN58, hong2019mina} (such as atrial fibrillation). Single-lead ECG records the heart's electrical activity from only one perspective, which can be seen as one partial view of a 12-lead ECG. A large gap in arrhythmia diagnosis still subsists between single-lead and multi-lead, indicating that the number of leads plays a key role in achieving better results (see Table. \ref{table2}). 

Recently, Knowledge Distillation methods are proposed to improve single-lead ECG's performance \cite{SEPAHVAND202264, hong2018rdpd}, where the teacher model was already developed through multi-lead ECG signals and the student model was developed through single-lead ECG signals. Since the student is educated under the supervision of the teacher to mimic the output space and inner feature or attention maps of its teacher, the internal feature maps to be processed must be strictly consistent. And this method can only diagnose ECG with single-label. It is noted that mutual information plays a key role in improving single-lead ECG's potential. So limiting consideration on the information relationship between single-lead ECG and multi-lead ECG, just forcing the same output of student with the teacher does not guarantee the robustness of the student.

In this paper, in order to efficiently stimulate the single-lead ECG signals' potential, we propose a novel and efficient architecture to transfer multi-view information of ECG for multi-disease label classification --- Multi-View Knowledge Transferring of ECG (MVKT-ECG). We adopt the teacher-student paradigm and extend it to a more general inter-lead knowledge transfer approach. Since mutual information is the key to improving single-lead ECG's ability, we formulate a new objective to maximize the mutual information of the feature representations between teacher and student --- Contrastive Lead-information Transferring (CLT) Loss. The CLT can maximize a lower bound to the mutual information between the teacher and student representations. Further, We modify traditional Knowledge Distillation to multi-label Knowledge Distillation to make it applicable to the context of multi-disease labels.  To verify the effectiveness of our method, in this paper, we conduct extensive experiments on two commonly used public datasets, PTB-XL and ICBEB2018. Experimental results demonstrate the effectiveness and robustness of multi-view knowledge transferring in the single-lead ECG classification task.
To sum up, the contributions of this paper include the following: 
\begin{itemize}
    \item We propose a more general and efficient framework MVKT-ECG to transfer multi-view information from multi-lead ECG signals to single-lead-based models.
    \item We explore the nature of inter-lead knowledge transfer and design a novel inter-lead information transferring objective—Contrastive Lead Transferring(CLT) Loss.
    \item We formulate multi-label Knowledge Distillation (MKD) and apply it to to fit the context of more general multi-disease label detection. 
\end{itemize}
\section{Related Work}
 \cite{hannun2019cardiologist} firstly proposed a comprehensive evaluation of an end-to-end deep neural network for ECG analysis across various diagnostic classes, demonstrating that an end-to-end deep learning approach can classify a broad range of distinct arrhythmias from single-lead ECGs with high diagnostic performance similar to that of cardiologists. Due to the limited information provided by single lead ECGs, all kinds of intelligent ECG equipment can only support a few types of detection at present. For example, the Apple Watch can only detect sinus rhythm(SR) and atrial fibrillation(AF) \cite{RN58}. And most academic research has also focused on that. Existing studies have achieved high detection accuracy in the detection of AF \cite{8331486}. However, limited by input information, the single-lead model performs poorly in classifying other arrhythmias, such as STD and PAC (Relevant experimental data will be presented in the Appendix).
\begin{figure*}[t]
\centering
\includegraphics[width=\textwidth]{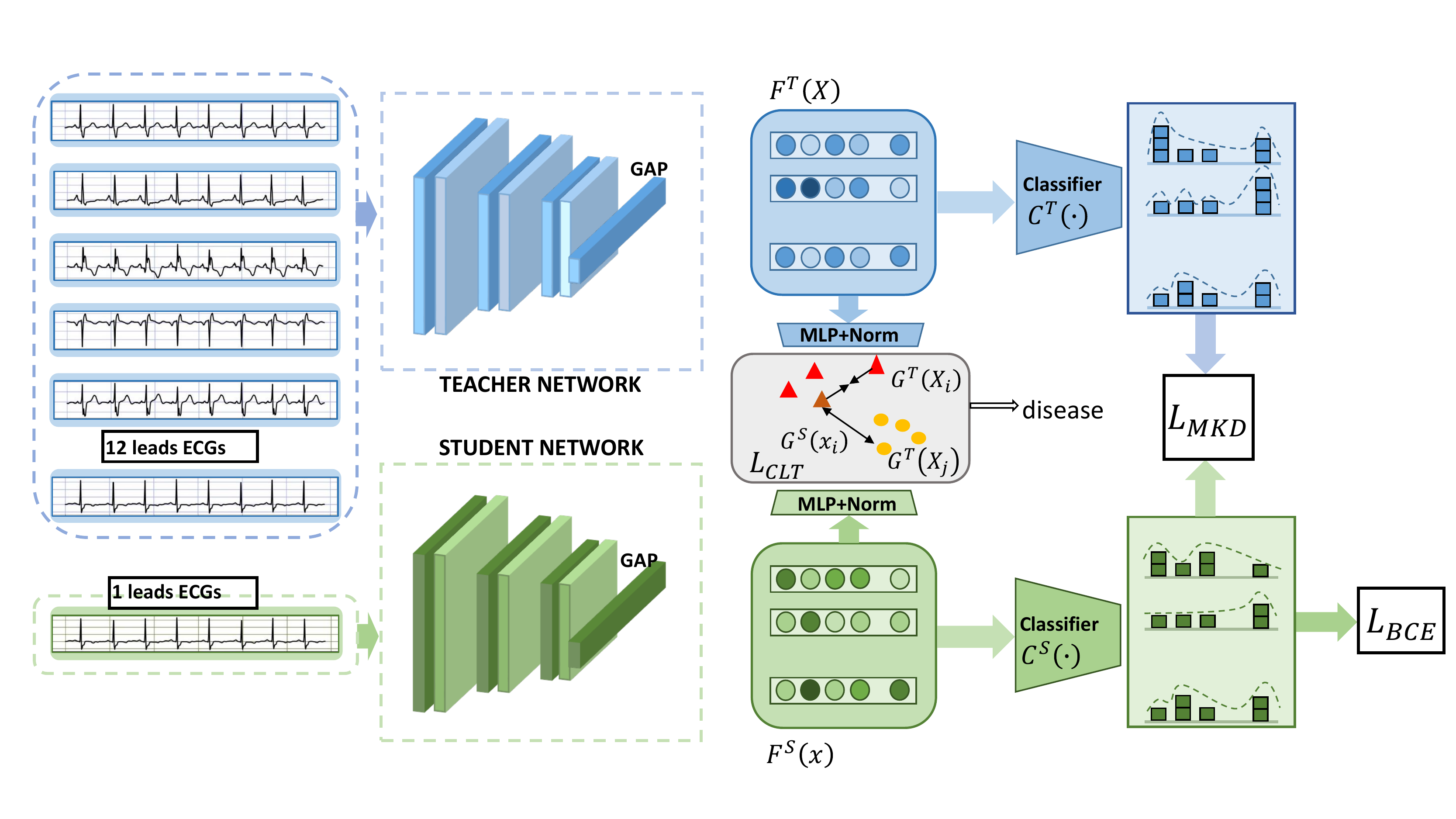} 
\caption{\textbf{An Overview of Multi-View Knowledge Transferring: the network using single-lead ECG signals is optimized by the network using 12-lead ECG signals.} The student network receives the multi-view Knowledge in two ways: (a) minimizing $\mathcal{L}_{\rm CLT}$ to improve the inter-lead disease mutual information (b) matching the logits of the teacher by minimizing the $\mathcal{L}_{\rm MKD}$; }
\label{fig2}
\vskip -0.2in
\end{figure*}

To address this problem of lacking ECG views, people proposed the method of restruction and synthesis of ECG. \cite{zhangallecg2019} restored another nine leads based on noisy three-lead signals through LSTM. \cite{pmlr-v139-golany21a} transformed the reconstruction of missing leads into solving the least squares problem by mapping the known ECG leads to the corresponding Koopman space and using the theory that the Koopman operator is linear. \cite{https://doi.org/10.48550/arxiv.2105.06293} proposed the concept of ECG panorama and the corresponding generation network Nef-Net and further proposed disease-aware synthesis method---ME-GAN, which attains panoptic electrocardio representations conditioned on heart diseases. These methods can restructure or synthesize multi-view ECG signals, but lacking evaluation on downstream classification tasks makes it difficult to guarantee the quality of the reconstructed signals to detect diseases. 

To achieve model compression of single-lead-based model and minimize the performance gap between the arrhythmia classification model with multi-lead ECG signals and the arrhythmia classification model with single-lead ECG signals, \cite{SEPAHVAND202264} proposed the teacher-student architecture. In order to deploy models in the poor-data environment without requiring direct access to multi-modal data acquired from a rich-data environment, \cite{hong2018rdpd} proposed a knowledge distillation (KD) method (RDPD) to enhance a predictive model trained on poor data using knowledge distilled from a high-complexity model trained on rich, private data. These methods use the traditional method of KD, forcing the student to imitate the output and inner features or attention of the teacher. It realizes the compression of the model through mature knowledge distillation loss and reduces the gap between single-lead ECG and 12-lead ECG. But this method must require the same dimension of the feature map between teacher and student, which makes the network structure lack flexibility. Different from this work, we explore the nature of inter-lead information transferring and formulate two new losses. In addition, our architecture can also deal with the muli-label-disease task, which is more general and efficient. 

\section{Method}
\subsection{Architecture}
Here we briefly describe the framework of our method and introduce the notations used in this paper. Given a 12 leads ECG signal $\boldsymbol{X}=\left[x^{(1)},x^{(2)},...,x^{(12)}\right] $ or a single lead ECG signal $\boldsymbol{x}=x^{(i)}$, where $\boldsymbol{X}\in \mathbb{R}^{12\times L}$ and $\boldsymbol{x}\in \mathbb{R}^{L}$. Our models are aimed to learn a function $\mathcal{F}_{\theta}(\boldsymbol{X})$, mapping a set of ECGs $\left[X_1, X_2,...,X_N\right] $ to a representative features space. Then the classifier $\mathcal{C}_{\theta}(\cdot)$ will project the above features into category space $\hat{Y} \in \mathbb{R}^{C}$. We consider the medical reality that sometimes patients with heart disease often have complex conditions, often multiple arrhythmias. So we consider the multi-label disease scenario, so the $sigmoid$ function was used as the final activation function. Here, we hope that this kind of representation can 1)minimize the information gap between single-lead ECGs and multi-lead ECGs signals and 2)be robust to the lead reduction in the single-lead ECGs. To achieve this, as shown in Figure \ref{fig2}, our proposed method is a two-stage procedure, as follows.\\
\textbf{First step }: we train the teacher network using standard 12 leads ECGs $\boldsymbol{X}=\left[x^{(1)},x^{(2)},...,x^{(12)}\right] $. The module $\mathcal{F}^T_{\theta}(\boldsymbol{X}): \mathbb{R}^{12\times L}\rightarrow \mathbb{R}^{D}$ will map each ECG signal $X_i$ to a fixed-size representation $d_i$ (in $xresnet1d101$ \cite{9190034}  $~D= 4096$). We train the teacher network-which will observe 12-lead ECG using the multi-label classification term --- Binary Cross Entropy Loss ($\mathcal{L}_{\rm BCE}$), which can be formulated as:
\begin{equation}\label{eqn:1}
\begin{aligned}
\mathcal{L}_{\rm BCE}=-\left[\boldsymbol{y}\log \hat{\boldsymbol{y}}+(1-\boldsymbol{y})\log(1-\hat{\boldsymbol{y}})  \right]
\end{aligned}
\end{equation}
where $\boldsymbol{y}$ represents the one-hot labels and $\hat{\boldsymbol{y}}$ represents the predicted classes probability.\\
\textbf{Second step}: we use the teacher network trained in the first step to instantiate the new student network, which only observes single-lead ECG, and the teacher's parameters are frozen. When we have trained the teacher model, we hope the representation ability of the network, which only observes single-lead ECGs, can be improved. In order to achieve this, we aim at the information knowledge we can gather from different leads, depicting the same heart condition under different views. When facing a 12-lead ECG classification task, one can often exploit lead viewpoints to provide a variety of appearances for a target disease. We want to teach a network to recover as much full-lead information just from a single-lead ECG. Although some diseases can not be inferred from single-lead ECG signals, what we want to do is to shorten the information gap between single-lead ECG and 12-lead ECG as far as possible, encouraging the student to focus more on key details to some particular diseases and maximum the mutual disease information between single-lead and 12-lead, further to reduce the incidence of misdiagnosis.  

Typically, due to the information gap between 12-lead ECG and single-lead ECG, the teacher network is always more powerful than the student network. This kind of asymmetry between the teacher and the student can produce a distillation objective different from the one due to the differences in model complexity. MVKT-ECG improves the mutual information between the feature representations of teacher and student by the Contrastive Lead-information Transferring (CLT) Loss. In the process of doing so, we also allow the student to imitate the teacher's output from the single-lead, which is a part of the 12-lead.  To do it, we refine the Knowledge Distillation loss \cite{hinton2015distilling} to a new knowledge distillation objective that can process multi-label disease problems.

\subsection{Contrastive Lead-information Transferring(CLT)}
To improve the inter-lead mutual information, we propose the disease-aware Contrast Lead-information Transferring(CLT) Loss, which can transfer useful disease information by maximizing the mutual information between single-lead ECG and multi-lead ECG.

As we all know, due to the lack of data dimensions, the disease information in the feature extracted by a neural network from single-lead ECG is less compared with 12-lead ECG, but it still contains part disease information. We refer to this information as the mutual information between single-lead ECG and 12-lead ECG. In another aspect, the absence of available data for single-lead ECG signals also tends to trigger a false diagnosis. For example, abnormal T wave morphology in any lead would be clinically diagnosed as a T wave change. In the case of abnormal T waves in other leads, judgment based on lead I alone is likely to lead to missed detection Fig \ref{fig1: sub_figure2}. We refer to this information that triggers a tendency to misdiagnosis as misleading information.

We know the lower limit of mutual information \cite{oord2018representation}:
\begin{equation}\label{eqn:2}
\begin{aligned}
& I (G^T(X),G^S(X))\\
&= \sum_{i,j}p(G^T(X_i),G^S(X_j)) \log \frac{p(G^T(X_i),G^S(X_j))}{p(G^T(X_i))(G^S(X_j))}\\
&\geq \log(N)+\mathbb{E}\left[\log \frac{f(G^T(X_i),G^S(X_j))}{f(G^T(X_i))(G^S(X_j))} \right]\\
&=  \log(N)+\mathbb{E}\left[\log \frac{exp\left[{(G^T(X_i) G^S(X_j))/\tau}\right]}{\sum exp\left[{(G^T(X_i)G^S(X_j))/\tau}\right]} \right]\\
&= \log(N) - \mathcal{L}_{\rm CLT}
\end{aligned}
\end{equation}
 The key idea of contrastive learning is learning a representation that is close for ``positive" pairs and pushing apart the representation between ``negative" pairs. In detail, for samples $X_i$ and $X_j$, MVKT-ECG shortens the student model and teacher model's representation of the same sample $G^S(x_i)$ and $G^T(X_i)$ and further the feature representation of the two models for different samples, such as $G^S(X_i)$ and $G^T(X_j)$. This can be achieved by minimizing:
\begin{equation}\label{eqn:3}
\begin{aligned}
&\mathcal{L}_{\rm CLT}=-\mathbb{E}\left[\log \frac{exp\left[{(G^T(X_i) G^S(X_j))/\tau}\right]}{\sum exp\left[{(G^T(X_i)G^S(X_j))/\tau}\right]} \right]\\
\end{aligned}
\end{equation}
\\
$N$ is the number of positive and negative sample pairs; the larger the better. Then a memory buffer is used to store the features of all training samples in this study. Since the teacher has observed multi-view ECGs, we believe that the different distances between ``positive" and ``negative" pairs yield a strong description of corresponding disease identities. During the training, $N$ samples are randomly selected to calculate the loss so that it can meet the demand of the model for the number of negative samples. The expression of each sample in the memory will carry on momentum updates according to the latest feature extracted from the network.
\subsection{Multi-label diseases Knowledge Distillation(MKD)}
The traditional KD focuses on single-label classification problems, so we propose MKD, a new knowledge distillation objective that can process multi-label disease problems.
Especially, we separate the output probability $p_i$ with $1-p_i$, and for each sample, we calculate the fore-and-aft softmax by temperature coefficient $\tau_{kd}$.
\begin{equation}\label{eqn:4}
\begin{aligned}
q_i = \frac{e^{p_i/\tau_{kd}}}{e^{p_i/\tau_{kd}}+e^{{(1-p_i)}/{\tau_{kd}}}}
\end{aligned}
\end{equation}

\begin{equation}\label{eqn:5}
\begin{aligned}
\mathcal{L}_{\rm MKD} =\tau^2_{kd} \sum_{i} {\rm KL} \left(q_i^T|| q_i^S\right)
\end{aligned}
\end{equation}
where $\tau_{kd}$ represents the temperature coefficient of multi-label disease knowledge distillation loss.
\subsection{Student Optimization}
In summary, the MVKT-ECG overall objective combines the distillation terms ($\mathcal{L}_{\rm MKD}$), the Contrastive  Lead-information Transferring terms($\mathcal{L}_{\rm CLT}$), and the ones optimized by hard label-$\mathcal{L}_{\rm BCE}$, which provide a higher conditional likelihood w.r.t. ground truth labels. To sum up, MVKT aims to boost single lead ECG's representation ability by the following optimization problem:
\begin{equation}\label{eqn:6}
\begin{aligned}
\operatorname{argmin}_{\theta_{S}} \mathcal{L}_{\rm MVKT}=\mathcal{L}_{\rm BCE}^{S}+\alpha \mathcal{L}_{\rm MKD}+\beta \mathcal{L}_{\rm CLT}
\end{aligned}
\end{equation}
where $\alpha$, $\beta$ are hyperparameters balancing the contributions of $\mathcal{L}_{\rm MKD}$ and $\mathcal{L}_{\rm CLT}$ to the total loss $\mathcal{L}_{\rm MVKT}$. Regarding the student initialization, we found that the CLECG's self-supervised strategy \cite{9543620} is beneficial, which will be mentioned in Sect. 5.5.
\begin{table*}[t]
\centering
\renewcommand\arraystretch{1.4}
\caption{Comparision MVKT's performance with State-of-the-Art single-lead ECG interpretation methods on public datasets. Since KD loss is only applicable to single-label classification, we use the original output KL-div when computing KD loss on these two data sets.}
\label{table1}
\vskip 0.15in
\resizebox*{0.99\textwidth}{0.26\textheight}{
    \begin{tabular}{c|ccc|ccc|ccc}
    \toprule[1.5pt]
  
    \multicolumn{1}{c|}{\multirow{2}{*}{Method}} & \multicolumn{3}{c|}{ICBEB2018}                                   & \multicolumn{3}{c|}{PTBXL.subdiagnostic}                       & \multicolumn{3}{c}{PTBXL.superdiagostic}                      \\ 
    \cline{2-10} 
    \multicolumn{1}{c|}{} & \multicolumn{1}{c}{AUC}  & \multicolumn{1}{c}{ACC}  & F1-score & \multicolumn{1}{c}{AUC} & \multicolumn{1}{c}{ACC} & F1-score & \multicolumn{1}{c}{AUC} & \multicolumn{1}{c}{ACC} & F1-score \\ \hline

    resnet1d\_Liu & \multicolumn{1}{c}{90.2} & \multicolumn{1}{c}{92.2} & 65.6     & \multicolumn{1}{c}{84.9}    & \multicolumn{1}{c}{88.5}    &     48.8     & \multicolumn{1}{c}{82.6}    & \multicolumn{1}{c}{80.2}    & 60.9         \\ 

    DenseCNN\_wang   & \multicolumn{1}{c}{88.7} & \multicolumn{1}{c}{90.6} & 61.7     & \multicolumn{1}{c}{84.5}    & \multicolumn{1}{c}{88.3}    &    47.9      & \multicolumn{1}{c}{82.7}    & \multicolumn{1}{c}{81.1}    &  61.7        \\ 

    SEresnet18     & \multicolumn{1}{c}{93.5} & \multicolumn{1}{c}{94.4} & 72.5     & \multicolumn{1}{c}{85.0}    & \multicolumn{1}{c}{88.8}    &     48.5     & \multicolumn{1}{c}{82.7}    & \multicolumn{1}{c}{81.3}    & 61.6         \\ 
    SEresnet34     & \multicolumn{1}{c}{93.7} & \multicolumn{1}{c}{94.4} & 73.1     & \multicolumn{1}{c}{84.4}    & \multicolumn{1}{c}{89.3}    &    47.7      & \multicolumn{1}{c}{82.9}    & \multicolumn{1}{c}{81.8}    & 62.3         \\ 
    CNN\_Hannun     & \multicolumn{1}{c}{94.0} & \multicolumn{1}{c}{94.7} & 74.6     & \multicolumn{1}{c}{84.9}    & \multicolumn{1}{c}{89.2}    &      48.9    & \multicolumn{1}{c}{83.1}    & \multicolumn{1}{c}{80.8}    &61.4         \\ 
    \hline 
    KD+FitNet & \multicolumn{1}{c}{94.0}&\multicolumn{1}{c}{94.8}&74.8 &\multicolumn{1}{c}{83.6} &\multicolumn{1}{c}{87.2} &46.4 & \multicolumn{1}{c}{83.8}& \multicolumn{1}{c}{76.5} &  60.1 \\

     KD+AT & \multicolumn{1}{c}{92.8}&\multicolumn{1}{c}{93.8}&71.4 &\multicolumn{1}{c}{82.0} &\multicolumn{1}{c}{86.1} &44.2 & \multicolumn{1}{c}{80.0}& \multicolumn{1}{c}{78.5} &  59.8 \\
     \hline

    MVKT-ECG     & \multicolumn{1}{c}{\textbf{95.7}} & \multicolumn{1}{c}{\textbf{95.7}} & \textbf{78.0}     & \multicolumn{1}{c}{\textbf{86.1}}    & \multicolumn{1}{c}{\textbf{88.3}}    &     \textbf{58.3}     & \multicolumn{1}{c}{\textbf{84.3}}    & \multicolumn{1}{c}{\textbf{82.2}} & \textbf{62.6} \\ 

    \toprule 
   \end{tabular} 
}
\vskip -0.1in
\end{table*}
\section{Dataset}
To build robust and efficient single-lead ECG interpretation models in a multi-label disease context. So we conduct experiments on two multi-label freely accessible datasets.\\
\noindent\textbf{ICBEB2018} dataset\cite{AnOpenAccessDatabase} contains 6,877 12-lead ECG recordings, each ranging in length from 6 to 60 seconds. The dataset considered one normal and eight abnormal arrhythmia categories, including AF, I-AVB, LBBB, PAC, PVC, RBBB, STD, and STE. Each record may have more than one label. This paper follows the processing method in \cite{AnOpenAccessDatabase} to divide the dataset into ten folds by stratified sampling, and the original label distribution is maintained in each fold. Among them, the first eight were used as training sets, while the 9th and 10th were used as verification and test sets, respectively. Then, we preprocess them to get equal-length data, and the length $n = 10000$.

\noindent\textbf{PTB-XL} dataset\cite{wagner2020ptb} is the to-date largest freely accessible clinical 12-lead ECG-waveform dataset comprising 21,837 records from 18,885 patients of 10 seconds long. This work uses all signals down-sampled at 100 Hz as the labeled data source. This dataset provides rich multi-level annotations, which, in terms of diagnosis, include superclasses of 5 classes and subclasses of 24 classes. We followed the Settings in \cite{wagner2020ptb} to divide the dataset into ten folds of class-balanced, with the first eight folds as the training set and the 9th and 10th folds as the verification set and test set, respectively. The signal length of each sample is 1,000 points.

\section{Experiments}
\textbf{Evaluation Metrics.} In the followings, we report performance in terms of Area under the Receiver Operating Characteristic (ROC-AUC) and the F1 score. 
\subsection{Experimental setups}
We implemented the MVKT algorithm in the framework of Pytorch. During the training, the model was trained according to the two-stage teacher-student training process, and the 12-lead data was first used to train the teacher model. All the teacher networks are trained for 100 epochs using Adam \cite{https://doi.org/10.48550/arxiv.1412.6980}. During the information transferring period, the parameters of the teacher model are frozen, and only the student model is updated. We feed 12-lead ECG signals to the teacher and single-lead ECGs to the student. In the process of information transferring, the temperature coefficient of knowledge distillation $\tau_{KD}$ is set as 1.5 because the dataset we used contains not many categories and the network prediction scores for each category are not concentrated. The temperature coefficient in the loss of comparative representation $\tau$ is set as 0.07, and the dimension of teacher and student representation $d$ is 128 in all structures. The size of the training batch is 32. In each training round, 1024 samples are randomly selected from the memory storage area to calculate the loss-$\mathcal{L}_{\rm CLT}$. In the concrete realization, the negative samples are extracted with the student output and the teacher output as the anchor points, respectively. The loss is composed of two symmetric parts. In the concrete realization, the negative samples are extracted with the student output and the teacher output as the anchor points, respectively. The loss is composed of two symmetric parts.
\subsection{Comparision with State-Of-The-Art Single Lead ECG's Classification Methods}
Table \ref{table1} reports a thorough comparison with current state-of-the-art (SOTA) methods across datasets. In the Physionet Challenge 2017 competition \cite{8331486}, Hannun’s CNN based model \cite{hannun2019cardiologist} achieved the best score. The DenseCNN\_wang \cite{10.1007/978-3-030-33327-0_9}, and resnet1d\_Liu \cite{10.1007/978-3-030-33327-0_11} achieved the top performance in China ECG AI Contest 2019 competition. \cite{9344414} proposed a large kernel size model SEresnet\_Zhao based on SE-block \cite{Hu_2018_CVPR}, achieving second place in the PhysioNet 2020 competition \cite{alday2020classification}. \cite{SEPAHVAND202264} (KD+FitNet) and \cite{hong2018rdpd} (KD + attention) also use the distillation idea to bridge the gap between the model with multi-lead ECG signals and single-lead ECG signals by trying to make the student's output and inner features the same as the teacher's. Compared with KD, Our proposed method performs better and outperforms other SOTA single-lead models. This result is fully consistent with our goal of robustness when providing only a single lead ECG as a query. 

\subsection{MVKT-ECG on different backbones}

We indicate the baseline and teacher model with the name of the backbone and append ``MVKT" for the model after the MVKT (e.g. ResMVKT34). We first benchmark 5 SOTA models' performance when they observe 12-lead ECGs and single-lead ECGs, respectively. The 5 models: CNN\_Hannun \cite{hannun2019cardiologist}, ResNet1d34 \cite{He_2016_CVPR}, ResNet1d\_wang \cite{wang2017time}, inception1d \cite{ismail2020inceptiontime}, XresNet1d101 \cite{strodthoff2020deep} are set as our backbones. The backbones' performance on single-lead ECGs is set as our baseline. Table \ref{table2} reports the comparison for different backbones. \cref{table4} and \cref{table5} demonstrates detail promotion of each disease. In detail, the student XresMVKT1d101 outperforms its baseline observably (3.2 \% on ICBEB2018, 1.3 \% on PTB.super, and 1.4 \% on PTB.sub). Based on this, we draw the following conclusions: 1) according to the objective which the student seeks to optimize, our method can gain significant improvement when single-lead ECGs are available; 2) compared with the baseline, the students do gain improvement, but it is hard for them to exceed the teacher's performance. As far as we can tell, the single-lead ECGs contain only a fraction of the 12-lead ECGs information that the MVKT can elicit to the greatest extent, but some pieces of information which single-lead ECG drops can never be recovered.
\begin{table}[t]
\centering
\renewcommand\arraystretch{1.4}
\caption{MVKT's results on different datasets, settings, and architectures.}
\label{table2}
\vskip 0.11in
\resizebox{0.95\columnwidth}{!}{
    \begin{tabular}{ccccc}
    \toprule[1.5pt]
    \multicolumn{1}{c}{\multirow{2}*{}}&\multicolumn{1}{c}{\multirow{2}*{\small{backbone}}}&\multicolumn{3}{c}{\multirow{1}*{\small{AUC (\%)}}}\\
    \cline{3-5}
    &&\small{PTB.sub}&\small{PTB.super}&\small{ICBEB2018}\\
    \hline
    Teacher&CNN\_Hannun&92.3&92.0&97.0\\
    \multicolumn{1}{c}{\multirow{2}*{1 lead}}&CNN\_Hannun&84.8&83.3&94.0\\
    &CNN\_Hannun\_MVKT&\textbf{85.6}&\textbf{83.4}&\textbf{95.7}\\
    \hline
    Teacher&ResNet1d34&92.3&92.8&94.5\\
    \multicolumn{1}{c}{\multirow{2}*{1 lead}}&ResNet1d34&84.5&82.6&90.7\\
    &ResMVKT34&\textbf{85.2}&\textbf{83.8}&\textbf{93.9}\\
    \hline
    Teacher&ResNet1d\_wang&92.7&92.1&91.0\\
    \multicolumn{1}{c}{\multirow{2}*{1 lead}}&ResNet1d\_wang&84.9&83.1&87.3\\
    &ResMVKT\_wang&\textbf{85.8}&\textbf{84.0}&\textbf{89.3}\\
    \hline
    Teacher&Inception1d&93.2&92.4&93.8\\
    \multicolumn{1}{c}{\multirow{2}*{1 lead}}&Inception1d&85.2&83.2&88.1\\
    &InceptionMVKT&\textbf{86.1}&\textbf{84.3}&\textbf{90.7}\\
    \hline
    Teacher&XresNet1d101&92.1&93.7&95.3\\
    \multicolumn{1}{c}{\multirow{2}*{1 lead}}&XresNet1d101&83.9&82.7&92.6\\
    &XresMVKT101&\textbf{85.3}&\textbf{84.0}&\textbf{94.6}\\
    \toprule[1.5pt]
    \end{tabular}
    }
\vskip -0.1in
\end{table}

\begin{figure}[H]
\centering
\includegraphics[width=0.99\columnwidth]{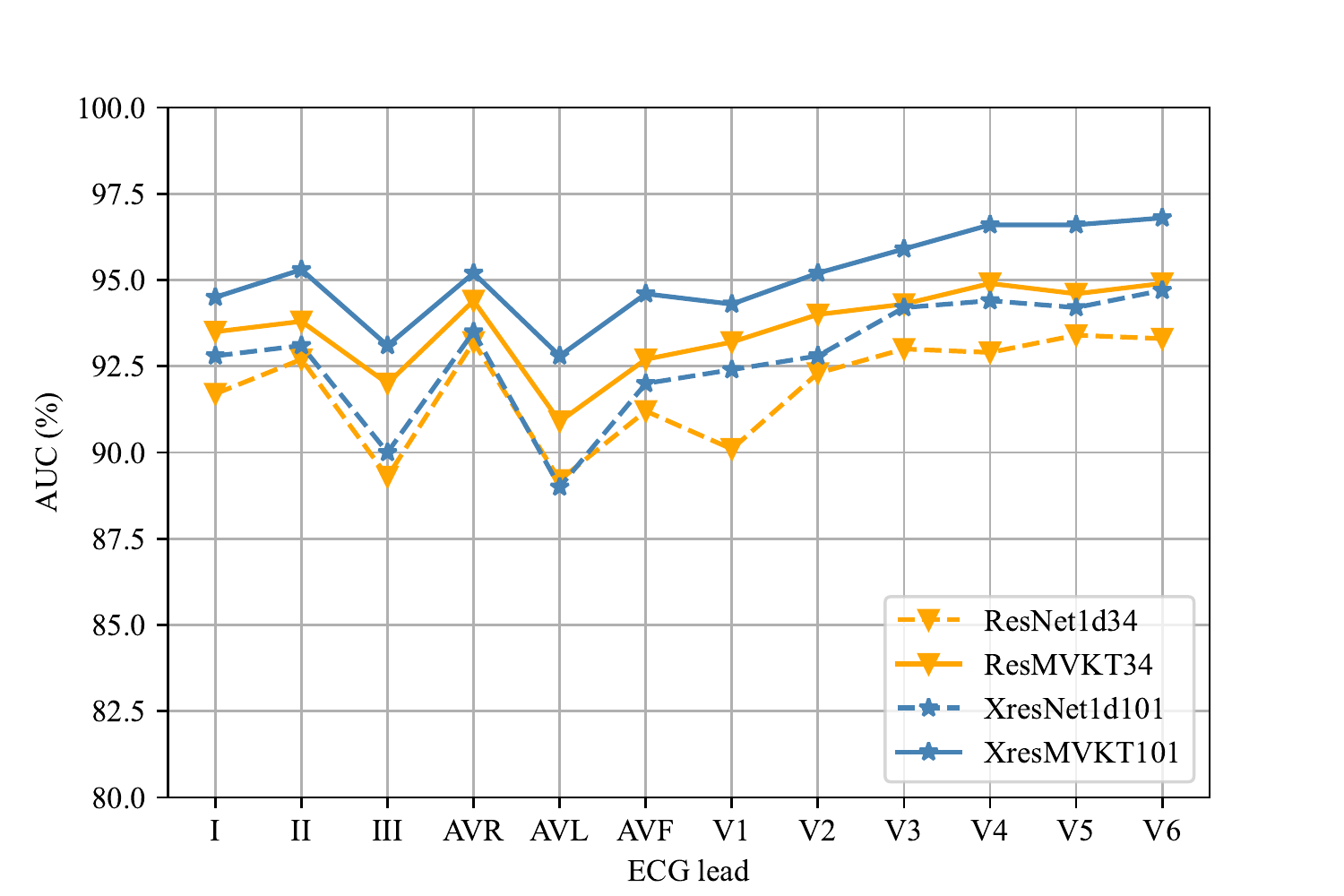} 
\caption{Performance (AUC) at evaluation time when changing the lead of ECG in ICBEB2018 dataset.}
\label{fig3}
\vskip -0.2in
\end{figure}

As additional proof, plots from Figure \ref{fig3} draw a comparison between models before and after information transferring. MVKT-ECG improves the performance considerably on ICBEB2018 dataset for each selected lead. Surprisingly, the lead II, AVR and limb lead (always v3--v6) always perform better. Surprisingly, MVKT-ECG can make lighter single-lead networks superior to complex models several times deeper: as an example, ResMVKT34 scores better than even XresNet101 on ICBEB2018, regardless of the selected lead of ECGs.


\subsection{Analysis on $\mathtt{\textbf{MVKT-ECG}}$}
\textbf{The impact of Loss Terms.}\\
We conducted a thorough ablation study for the loss terms. Without loss of generality, we focus our analysis on XresMVKT101 on the ICBEB2018 dataset. Table \ref{table3} reports the ablation’s result. In the three losses, $\mathcal{L}_{\rm CLT}$ plays an important role. Although we only use the hard label and $\mathcal{L}_{\rm CLT}$, it can also achieve a significant result, which also demonstrates our proposed role of information transferring between leads. And as expected, the greatest performance of AUC is obtained with all the losses.
\begin{table}[h]
\centering
\caption{Ablation study in terms of the impact of each loss term.}
\label{table3}
\vskip 0.11in
\renewcommand\arraystretch{1.4}
\resizebox{\columnwidth}{!}{
\begin{tabular}{ccccc}
\toprule[1.5pt]

&$\mathcal{L}_{\rm BCE}$&$\mathcal{L}_{\rm MKD}$&$\mathcal{L}_{\rm CLT}$&{AUC (\%)}\\
\hline
{XresNet101(Teacher)}&&&&{{95.3}}\\
\hline
\multicolumn{1}{c}{\multirow{4}{*}{\makecell{\small XresMVKT101\\\small (Student)}}}&\checkmark&&&{88.2}\\
&\checkmark&\checkmark&&{93.7}\\
&\checkmark&&\checkmark&{93.9}\\
&\checkmark&\checkmark&\checkmark&\textbf{94.4}\\

\toprule[1.5pt]
\end{tabular}
}
\vskip -0.1in
\end{table}

\begin{figure*}[t]
\centering
\includegraphics[width=0.9\textwidth]{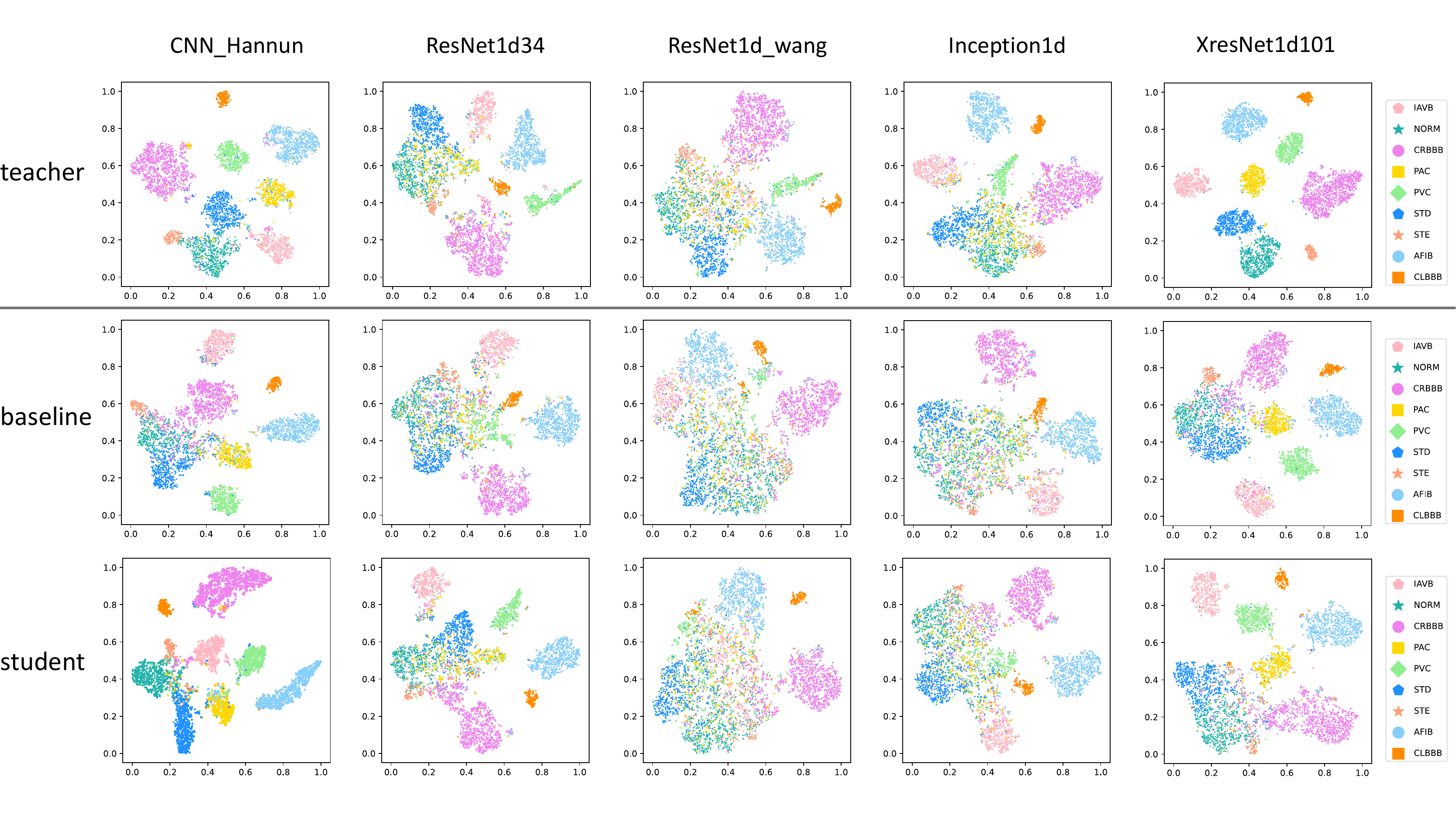} 
\caption{\textbf{Feature visualization.} The first line, second and third lines are the feature distribution of teacher, baseline, and student, respectively.}
\label{fig4}
\vskip -0.2in
\end{figure*}

\noindent\textbf{Visualization of the classes distribution.}\\
To visually assess the differences between the baseline and student, we use the T-SNE graphs to highlight the feature distribution. Figure \ref{fig4} depicts the impact of MVKT between different backbones on ICBEB2018. As we can see, the distribution of features of normal (Norm), atrial premature beat (PAC), ST-segment depression (STD), and ST-segment elevation (STE) in the baseline network are confused to some extent. After MVKT, although the four categories are still relatively concentrated in the same area, the feature distribution of the distilled model in all nine categories became denser. As for the issue of category overlap, the status is greatly improved, especially on CNN\_Hannun, ResNet1d34, and XresNet1d101 backbones. (We can see the improvement of each disease on ResNet1d\_wang and Inception1d from \cref{table4}). This suggests that the information conveyed by other leads can make up for the lack of single leads for various arrhythmias. \\

\noindent\textbf{Commonality.}\\
Faced with the shortage of ECG datasets, many self-supervised learning methods are proposed to take advantage of unlabeled data. As for the student network's initialization, \cite{9543620} introduced a SOTA unsupervised pre-training program --- CLECG to mine adequate information from unlabeled data. During the pre-training, CLECG encourages the representations of different augmented views of the same signal to be similar and increases the distance between representations of augmented views from the different signals.

We combine MVKT-ECG with CLECG. As PTB-XL dataset is the to-date largest freely accessible clinical ECG-waveform dataset, we firstly pre-train the network with CLECG on PTB-XL dataset, then using MVKT-ECG to finetune. The different proportions (eighth, quarter, half, all) of ICBEB2018 training set data were used in funtune procedure. AUC indicators of random initialization and self-supervised pre-training were compared and the results are shown in Figure \ref{fig5}. 

It can be seen that at different data sizes, MVKT-ECG can improve performance by 1\%-2\% compared to the baseline model and even surpass the performance of the baseline model trained with total data when using only half of the data. On this basis, using the parameters obtained from external data of CLECG can further improve the performance of the student model, and the AUC metric increases by 0.9\% when using all training data, which shows that our method has practical application in combination with other methods. Moreover, this result suggests that the student model can improve performance by accepting knowledge transferred both from unlabeled external data and other leads.
\begin{figure}[h]
\centering
\includegraphics[width=0.95\columnwidth]{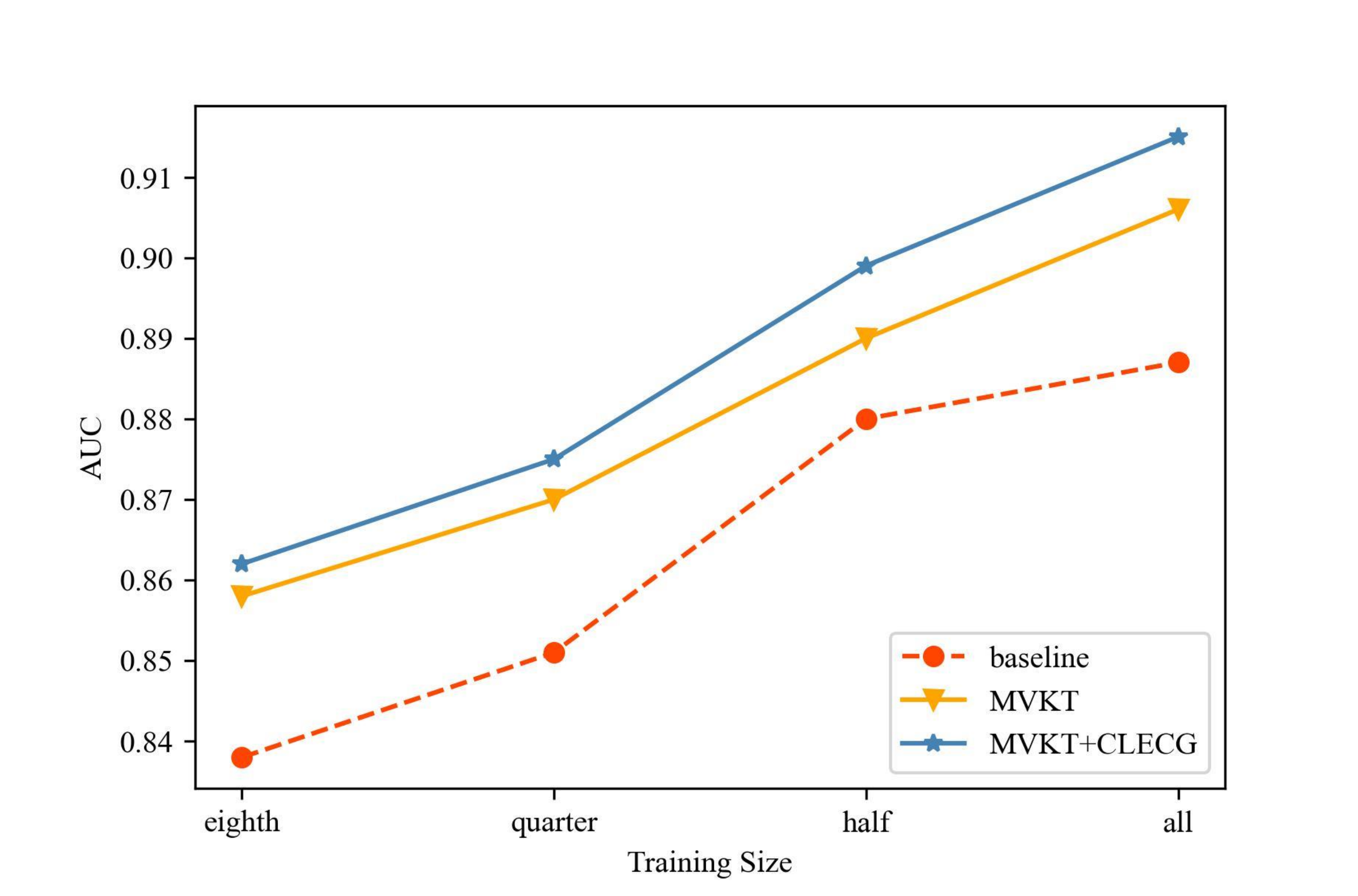} 
\caption{\textbf{Combine MVKT-ECG with CLECG.} The pre-training procedure is based on the PTB-XL dataset, and the finetune procedure is based on the ICBEB2018 dataset.}
\label{fig5}
\end{figure}

\section{Conclusion}
An effective ECG smart device recognition algorithm requires higher recognition accuracy and robustness for single-lead ECG signals. To achieve this goal, we propose MVKT-ECG. This is teacher-student information transferring approach in which the student observes only a single lead of the input ECG signals. This strategy encourages students to find better representations and to be closer to the teacher in performance through the knowledge transferring of the 12-lead ECG processing network. Notably, MVKT-ECG shows robustness in different datasets and different backbones. Experimental results show that the proposed algorithm is insensitive to the classification granularity and specific categories of the datasets, and the accuracy of the student model in the detection of multiple arrhythmias is greatly improved. The visualization results also show that multi-view knowledge distillation can guide the student model to simulate the expression of the teacher model, which makes up for the deficiency of single-lead ECGs in differentiating some specific categories.
\newpage
\nocite{langley00}

\bibliography{example_paper}

\begin{thebibliography}{34}
\providecommand{\natexlab}[1]{#1}
\providecommand{\url}[1]{\texttt{#1}}
\expandafter\ifx\csname urlstyle\endcsname\relax
  \providecommand{\doi}[1]{doi: #1}\else
  \providecommand{\doi}{doi: \begingroup \urlstyle{rm}\Url}\fi

\bibitem[Alday et~al.(2020)Alday, Gu, Shah, Robichaux, Wong, Liu, Liu, Rad,
  Elola, Seyedi, et~al.]{alday2020classification}
Alday, E. A.~P., Gu, A., Shah, A.~J., Robichaux, C., Wong, A.-K.~I., Liu, C.,
  Liu, F., Rad, A.~B., Elola, A., Seyedi, S., et~al.
\newblock Classification of 12-lead ecgs: the physionet/computing in cardiology
  challenge 2020.
\newblock \emph{Physiological measurement}, 41\penalty0 (12):\penalty0 124003,
  2020.

\bibitem[Chen et~al.(2021{\natexlab{a}})Chen, Wang, Zhang, Zhang, and
  Yang]{9543620}
Chen, H., Wang, G., Zhang, G., Zhang, P., and Yang, H.
\newblock Clecg: A novel contrastive learning framework for electrocardiogram
  arrhythmia classification.
\newblock \emph{IEEE Signal Processing Letters}, 28:\penalty0 1993--1997,
  2021{\natexlab{a}}.
\newblock \doi{10.1109/LSP.2021.3114119}.

\bibitem[Chen et~al.(2021{\natexlab{b}})Chen, Zheng, Yu, Chen, and
  Wu]{https://doi.org/10.48550/arxiv.2105.06293}
Chen, J., Zheng, X., Yu, H., Chen, D.~Z., and Wu, J.
\newblock Electrocardio panorama: Synthesizing new ecg views with
  self-supervision, 2021{\natexlab{b}}.
\newblock URL \url{https://arxiv.org/abs/2105.06293}.

\bibitem[Chen et~al.(2022)Chen, Liao, Wei, Ying, Chen, and
  Wu]{pmlr-v162-chen22n}
Chen, J., Liao, K., Wei, K., Ying, H., Chen, D.~Z., and Wu, J.
\newblock {ME}-{GAN}: Learning panoptic electrocardio representations for
  multi-view {ECG} synthesis conditioned on heart diseases.
\newblock In Chaudhuri, K., Jegelka, S., Song, L., Szepesvari, C., Niu, G., and
  Sabato, S. (eds.), \emph{Proceedings of the 39th International Conference on
  Machine Learning}, volume 162 of \emph{Proceedings of Machine Learning
  Research}, pp.\  3360--3370. PMLR, 17--23 Jul 2022.
\newblock URL \url{https://proceedings.mlr.press/v162/chen22n.html}.

\bibitem[Clifford et~al.(2017)Clifford, Liu, Moody, Lehman, Silva, Li, Johnson,
  and Mark]{8331486}
Clifford, G.~D., Liu, C., Moody, B., Lehman, L.-w.~H., Silva, I., Li, Q.,
  Johnson, A.~E., and Mark, R.~G.
\newblock Af classification from a short single lead ecg recording: The
  physionet/computing in cardiology challenge 2017.
\newblock In \emph{2017 Computing in Cardiology (CinC)}, pp.\  1--4, 2017.
\newblock \doi{10.22489/CinC.2017.065-469}.

\bibitem[Golany et~al.(2021)Golany, Radinsky, Freedman, and
  Minha]{pmlr-v139-golany21a}
Golany, T., Radinsky, K., Freedman, D., and Minha, S.
\newblock 12-lead ecg reconstruction via koopman operators.
\newblock In Meila, M. and Zhang, T. (eds.), \emph{Proceedings of the 38th
  International Conference on Machine Learning}, volume 139 of
  \emph{Proceedings of Machine Learning Research}, pp.\  3745--3754. PMLR,
  18--24 Jul 2021.
\newblock URL \url{https://proceedings.mlr.press/v139/golany21a.html}.

\bibitem[Hannun et~al.(2019)Hannun, Rajpurkar, Haghpanahi, Tison, Bourn,
  Turakhia, and Ng]{hannun2019cardiologist}
Hannun, A.~Y., Rajpurkar, P., Haghpanahi, M., Tison, G.~H., Bourn, C.,
  Turakhia, M.~P., and Ng, A.~Y.
\newblock Cardiologist-level arrhythmia detection and classification in
  ambulatory electrocardiograms using a deep neural network.
\newblock \emph{Nature medicine}, 25\penalty0 (1):\penalty0 65--69, 2019.

\bibitem[He et~al.(2016)He, Zhang, Ren, and Sun]{He_2016_CVPR}
He, K., Zhang, X., Ren, S., and Sun, J.
\newblock Deep residual learning for image recognition.
\newblock In \emph{Proceedings of the IEEE Conference on Computer Vision and
  Pattern Recognition (CVPR)}, June 2016.

\bibitem[Hinton et~al.(2015)Hinton, Vinyals, Dean,
  et~al.]{hinton2015distilling}
Hinton, G., Vinyals, O., Dean, J., et~al.
\newblock Distilling the knowledge in a neural network.
\newblock \emph{arXiv preprint arXiv:1503.02531}, 2\penalty0 (7), 2015.

\bibitem[Holst et~al.(1999)Holst, Ohlsson, Peterson, and
  Edenbrandt]{holst1999confident}
Holst, H., Ohlsson, M., Peterson, C., and Edenbrandt, L.
\newblock A confident decision support system for interpreting
  electrocardiograms.
\newblock \emph{Clinical Physiology}, 19\penalty0 (5):\penalty0 410--418, 1999.

\bibitem[Hong et~al.(2018)Hong, Xiao, Hoang, Ma, Li, and Sun]{hong2018rdpd}
Hong, S., Xiao, C., Hoang, T.~N., Ma, T., Li, H., and Sun, J.
\newblock Rdpd: rich data helps poor data via imitation.
\newblock \emph{arXiv preprint arXiv:1809.01921}, 2018.

\bibitem[Hong et~al.(2019)Hong, Xiao, Ma, Li, and Sun]{hong2019mina}
Hong, S., Xiao, C., Ma, T., Li, H., and Sun, J.
\newblock Mina: multilevel knowledge-guided attention for modeling
  electrocardiography signals.
\newblock \emph{arXiv preprint arXiv:1905.11333}, 2019.

\bibitem[Hu et~al.(2018)Hu, Shen, and Sun]{Hu_2018_CVPR}
Hu, J., Shen, L., and Sun, G.
\newblock Squeeze-and-excitation networks.
\newblock In \emph{Proceedings of the IEEE Conference on Computer Vision and
  Pattern Recognition (CVPR)}, June 2018.

\bibitem[Ismail~Fawaz et~al.(2020)Ismail~Fawaz, Lucas, Forestier, Pelletier,
  Schmidt, Weber, Webb, Idoumghar, Muller, and
  Petitjean]{ismail2020inceptiontime}
Ismail~Fawaz, H., Lucas, B., Forestier, G., Pelletier, C., Schmidt, D.~F.,
  Weber, J., Webb, G.~I., Idoumghar, L., Muller, P.-A., and Petitjean, F.
\newblock Inceptiontime: Finding alexnet for time series classification.
\newblock \emph{Data Mining and Knowledge Discovery}, 34\penalty0 (6):\penalty0
  1936--1962, 2020.

\bibitem[Kingma \& Ba(2014)Kingma and
  Ba]{https://doi.org/10.48550/arxiv.1412.6980}
Kingma, D.~P. and Ba, J.
\newblock Adam: A method for stochastic optimization, 2014.
\newblock URL \url{https://arxiv.org/abs/1412.6980}.

\bibitem[Kiranyaz et~al.(2016)Kiranyaz, Ince, and Gabbouj]{7202837}
Kiranyaz, S., Ince, T., and Gabbouj, M.
\newblock Real-time patient-specific ecg classification by 1-d convolutional
  neural networks.
\newblock \emph{IEEE Transactions on Biomedical Engineering}, 63\penalty0
  (3):\penalty0 664--675, 2016.
\newblock \doi{10.1109/TBME.2015.2468589}.

\bibitem[Kiyasseh et~al.(2021)Kiyasseh, Zhu, and Clifton]{kiyasseh2021clocs}
Kiyasseh, D., Zhu, T., and Clifton, D.~A.
\newblock Clocs: Contrastive learning of cardiac signals across space, time,
  and patients.
\newblock In \emph{International Conference on Machine Learning}, pp.\
  5606--5615. PMLR, 2021.

\bibitem[Liu et~al.(2018)Liu, Liu, Zhao, Zhang, Wu, Xu, Liu, Ma, Wei, He, Li,
  and Ng]{AnOpenAccessDatabase}
Liu, F., Liu, C., Zhao, L., Zhang, X., Wu, X., Xu, X., Liu, Y., Ma, C., Wei,
  S., He, Z., Li, J., and Ng, E.
\newblock An open access database for evaluating the algorithms of
  electrocardiogram rhythm and morphology abnormality detection.
\newblock \emph{Journal of Medical Imaging and Health Informatics}, 8:\penalty0
  1368--1373, 09 2018.
\newblock \doi{10.1166/jmihi.2018.2442}.

\bibitem[Liu et~al.(2019)Liu, He, Wang, Li, Sun, Zhao, and
  Zhang]{10.1007/978-3-030-33327-0_11}
Liu, Y., He, R., Wang, K., Li, Q., Sun, Q., Zhao, N., and Zhang, H.
\newblock Automatic detection of ecg abnormalities by using an ensemble of deep
  residual networks with attention.
\newblock In Liao, H., Balocco, S., Wang, G., Zhang, F., Liu, Y., Ding, Z.,
  Duong, L., Phellan, R., Zahnd, G., Breininger, K., Albarqouni, S., Moriconi,
  S., Lee, S.-L., and Demirci, S. (eds.), \emph{Machine Learning and Medical
  Engineering for Cardiovascular Health and Intravascular Imaging and Computer
  Assisted Stenting}, pp.\  88--95, Cham, 2019. Springer International
  Publishing.
\newblock ISBN 978-3-030-33327-0.

\bibitem[Liu et~al.(2021)Liu, Li, Wang, Liu, He, Yuan, and Zhang]{bios11110453}
Liu, Y., Li, Q., Wang, K., Liu, J., He, R., Yuan, Y., and Zhang, H.
\newblock Automatic multi-label ecg classification with category imbalance and
  cost-sensitive thresholding.
\newblock \emph{Biosensors}, 11\penalty0 (11), 2021.
\newblock ISSN 2079-6374.
\newblock \doi{10.3390/bios11110453}.
\newblock URL \url{https://www.mdpi.com/2079-6374/11/11/453}.

\bibitem[Oord et~al.(2018)Oord, Li, and Vinyals]{oord2018representation}
Oord, A. v.~d., Li, Y., and Vinyals, O.
\newblock Representation learning with contrastive predictive coding.
\newblock \emph{arXiv preprint arXiv:1807.03748}, 2018.

\bibitem[Ribeiro et~al.(2020)Ribeiro, Ribeiro, Paix{\~a}o, Oliveira, Gomes,
  Canazart, Ferreira, Andersson, Macfarlane, Meira~Jr,
  et~al.]{ribeiro2020automatic}
Ribeiro, A.~H., Ribeiro, M.~H., Paix{\~a}o, G.~M., Oliveira, D.~M., Gomes,
  P.~R., Canazart, J.~A., Ferreira, M.~P., Andersson, C.~R., Macfarlane, P.~W.,
  Meira~Jr, W., et~al.
\newblock Automatic diagnosis of the 12-lead ecg using a deep neural network.
\newblock \emph{Nature communications}, 11\penalty0 (1):\penalty0 1--9, 2020.

\bibitem[Rizas et~al.(2022)Rizas, Freyer, Sappler, von Stülpnagel,
  Spielbichler, Krasniqi, Schreinlechner, Wenner, Theurl, Behroz, Eiffener,
  Klemm, Schneidewind, Zens, Dolejsi, Mansmann, Massberg, and Bauer]{RN58}
Rizas, K.~D., Freyer, L., Sappler, N., von Stülpnagel, L., Spielbichler, P.,
  Krasniqi, A., Schreinlechner, M., Wenner, F.~N., Theurl, F., Behroz, A.,
  Eiffener, E., Klemm, M.~P., Schneidewind, A., Zens, M., Dolejsi, T.,
  Mansmann, U., Massberg, S., and Bauer, A.
\newblock Smartphone-based screening for atrial fibrillation: a pragmatic
  randomized clinical trial.
\newblock \emph{Nature Medicine}, 28\penalty0 (9):\penalty0 1823--1830, 2022.
\newblock ISSN 1546-170X.
\newblock \doi{10.1038/s41591-022-01979-w}.
\newblock URL \url{https://doi.org/10.1038/s41591-022-01979-w}.

\bibitem[Sarkar \& Etemad(2020)Sarkar and Etemad]{9161416}
Sarkar, P. and Etemad, A.
\newblock Self-supervised ecg representation learning for emotion recognition.
\newblock \emph{IEEE Transactions on Affective Computing}, pp.\  1--1, 2020.
\newblock \doi{10.1109/TAFFC.2020.3014842}.

\bibitem[Sepahvand \& Abdali-Mohammadi(2022)Sepahvand and
  Abdali-Mohammadi]{SEPAHVAND202264}
Sepahvand, M. and Abdali-Mohammadi, F.
\newblock A novel method for reducing arrhythmia classification from 12-lead
  ecg signals to single-lead ecg with minimal loss of accuracy through
  teacher-student knowledge distillation.
\newblock \emph{Information Sciences}, 593:\penalty0 64--77, 2022.
\newblock ISSN 0020-0255.
\newblock \doi{https://doi.org/10.1016/j.ins.2022.01.030}.
\newblock URL
  \url{https://www.sciencedirect.com/science/article/pii/S0020025522000457}.

\bibitem[Strodthoff et~al.(2020)Strodthoff, Wagner, Schaeffter, and
  Samek]{strodthoff2020deep}
Strodthoff, N., Wagner, P., Schaeffter, T., and Samek, W.
\newblock Deep learning for ecg analysis: Benchmarks and insights from ptb-xl.
\newblock \emph{IEEE Journal of Biomedical and Health Informatics}, 25\penalty0
  (5):\penalty0 1519--1528, 2020.

\bibitem[Strodthoff et~al.(2021)Strodthoff, Wagner, Schaeffter, and
  Samek]{9190034}
Strodthoff, N., Wagner, P., Schaeffter, T., and Samek, W.
\newblock Deep learning for ecg analysis: Benchmarks and insights from ptb-xl.
\newblock \emph{IEEE Journal of Biomedical and Health Informatics}, 25\penalty0
  (5):\penalty0 1519--1528, 2021.
\newblock \doi{10.1109/JBHI.2020.3022989}.

\bibitem[Wagner et~al.(2020)Wagner, Strodthoff, Bousseljot, Kreiseler, Lunze,
  Samek, and Schaeffter]{wagner2020ptb}
Wagner, P., Strodthoff, N., Bousseljot, R.-D., Kreiseler, D., Lunze, F.~I.,
  Samek, W., and Schaeffter, T.
\newblock Ptb-xl, a large publicly available electrocardiography dataset.
\newblock \emph{Scientific data}, 7\penalty0 (1):\penalty0 1--15, 2020.

\bibitem[Wang et~al.(2019{\natexlab{a}})Wang, Yang, Tang, and
  Li]{10.1007/978-3-030-33327-0_9}
Wang, C., Yang, S., Tang, X., and Li, B.
\newblock A 12-lead ecg arrhythmia classification method based on 1d densely
  connected cnn.
\newblock In Liao, H., Balocco, S., Wang, G., Zhang, F., Liu, Y., Ding, Z.,
  Duong, L., Phellan, R., Zahnd, G., Breininger, K., Albarqouni, S., Moriconi,
  S., Lee, S.-L., and Demirci, S. (eds.), \emph{Machine Learning and Medical
  Engineering for Cardiovascular Health and Intravascular Imaging and Computer
  Assisted Stenting}, pp.\  72--79, Cham, 2019{\natexlab{a}}. Springer
  International Publishing.
\newblock ISBN 978-3-030-33327-0.

\bibitem[Wang et~al.(2019{\natexlab{b}})Wang, Zhang, Liu, Yang, Fu, Wang, and
  Zhang]{WANG2019523}
Wang, G., Zhang, C., Liu, Y., Yang, H., Fu, D., Wang, H., and Zhang, P.
\newblock A global and updatable ecg beat classification system based on
  recurrent neural networks and active learning.
\newblock \emph{Information Sciences}, 501:\penalty0 523--542,
  2019{\natexlab{b}}.
\newblock ISSN 0020-0255.
\newblock \doi{https://doi.org/10.1016/j.ins.2018.06.062}.
\newblock URL
  \url{https://www.sciencedirect.com/science/article/pii/S0020025518305115}.

\bibitem[Wang et~al.(2017)Wang, Yan, and Oates]{wang2017time}
Wang, Z., Yan, W., and Oates, T.
\newblock Time series classification from scratch with deep neural networks: A
  strong baseline.
\newblock In \emph{2017 International joint conference on neural networks
  (IJCNN)}, pp.\  1578--1585. IEEE, 2017.

\bibitem[Xia et~al.(2019)Xia, Sang, Guo, Ji, Han, Chen, Yang, and
  Meng]{10.1007/978-3-030-33327-0_10}
Xia, Z., Sang, Z., Guo, Y., Ji, W., Han, C., Chen, Y., Yang, S., and Meng, L.
\newblock Automatic multi-label classification in 12-lead ecgs using neural
  networks and characteristic points.
\newblock In Liao, H., Balocco, S., Wang, G., Zhang, F., Liu, Y., Ding, Z.,
  Duong, L., Phellan, R., Zahnd, G., Breininger, K., Albarqouni, S., Moriconi,
  S., Lee, S.-L., and Demirci, S. (eds.), \emph{Machine Learning and Medical
  Engineering for Cardiovascular Health and Intravascular Imaging and Computer
  Assisted Stenting}, pp.\  80--87, Cham, 2019. Springer International
  Publishing.
\newblock ISBN 978-3-030-33327-0.

\bibitem[Zhang \& Frick(2019)Zhang and Frick]{zhangallecg2019}
Zhang, Q. and Frick, K.
\newblock All-ecg: A least-number of leads ecg monitor for standard 12-lead ecg
  tracking during motion.
\newblock In \emph{2019 IEEE Healthcare Innovations and Point of Care
  Technologies, (HI-POCT)}, pp.\  103--106, 2019.
\newblock \doi{10.1109/HI-POCT45284.2019.8962742}.

\bibitem[Zhao et~al.(2020)Zhao, Fang, Relton, Yan, Liu, Li, Qin, and
  Wong]{9344414}
Zhao, Z., Fang, H., Relton, S.~D., Yan, R., Liu, Y., Li, Z., Qin, J., and Wong,
  D.~C.
\newblock Adaptive lead weighted resnet trained with different duration signals
  for classifying 12-lead ecgs.
\newblock In \emph{2020 Computing in Cardiology}, pp.\  1--4, 2020.
\newblock \doi{10.22489/CinC.2020.112}.

\end{thebibliography}
\bibliographystyle{icml2023}

\newpage
\appendix
\onecolumn
\section{Detailed experimental results.}
Here we report the F1\_score of each disease on different backbones. We also try to use different frameworks for teachers and students. We can see from the \cref{table4} and \cref{table5} that MVKT is beneficial in most cases for the diagnosis of any disease except for CNN\_Hannun and Resnet1d\_wang. This suggests that the MVKT operates in an environment with sufficient model capacity.
\begin{table}[ht]
\centering
\caption{MVKT’s results of different diseases on ICBEB2018 dataset}
\label{table4}
\vskip 0.1in
\resizebox{\textwidth}{!}{%
\begin{tabular}{ccccccccccccc}
\toprule
&\multirow{2}[2]{*}{backbone}&\multirow{2}[2]{*}{MVKT}&\multicolumn{10}{c}{F1\_score (\%)}\\
\cmidrule{4-13}
&&&NORM&AF&I-AVB&LBBB&RBBB&PAC&PVC&STD&STE&Average\\
\midrule[1.5pt]
Teacher&CNN\_Hannun&&75.8&92.0&87.0&88.0&91.7&71.3&86.7&74.0&68.1&81.6\\

\multirow{2}{*}{Student}&\multirow{2}{*}{CNN\_Hannun}&&65.2&95.2&84.8&91.3&85.0&66.7&86.5&66.0&30.8&74.6\\
&&$\surd$&\textbf{73.9}&\textbf{95.9}&\textbf{85.5}&87.5&\textbf{86.9}&\textbf{70.8}&82.2&\textbf{72.0}&\textbf{47.4}&\textbf{78.0}\\
\midrule[1.5pt]
Teacher&ResNet1d34&&73.4&91.7&85.7&88.4&91.0&35.7&82.2&75.5&61.9&76.2\\
\midrule
\multirow{2}{*}{Student}&\multirow{2}{*}{ResNet1d34}&&63.9&92.1&84.5&87.5&85.5&33.8&60.0&65.1&26.9&66.6\\
&&$\surd$&\textbf{65.6}&\textbf{93.3}&\textbf{86.3}&\textbf{90.5}&\textbf{85.5}&\textbf{40.3}&\textbf{79.1}&\textbf{60.2}&\textbf{34.3}&\textbf{70.6}\\

\midrule[1.5pt]
Teacher&Resnet1d\_wang&&65.9&90.5&55.8&82.6&90.5&22.9&77.2&75.7&57.1&68.7\\
\midrule
\multirow{2}{*}{Student}&\multirow{2}{*}{Resnet1d\_wang}&&60.2&82.7&67.7&84.0&84.5&26.3&36.1&60.0&30.4&59.1\\
&&$\surd$&{59.4}&\textbf{85.9}&\textbf{68.5}&\textbf{87.0}&{84.0}&24.7&\textbf{40.7}&\textbf{61.2}&26.3&\textbf{59.8}\\
\midrule[1.5pt]
Teacher&Inception1d&&70.6&92.3&85.9&84.0&90.6&39.4&81.8&76.7&61.9&95.9\\
\midrule
\multirow{2}{*}{Student}&\multirow{2}{*}{Inception1d}&&62.4&89.9&81.1&85.1&84.07&27.9&30.9&60.0&26.9&60.9\\
&&$\surd$&\textbf{67.0}&\textbf{91.1}&\textbf{81.3}&\textbf{85.7}&\textbf{86.5}&\textbf{33.7}&\textbf{56.0}&\textbf{63.0}&\textbf{36.7}&\textbf{66.8}\\
\midrule[1.5pt]
Teacher&XresNet1d101&&73.7&95.2&85.3&88.9&92.5&60.2&89.6&77.0&66.7&81.0\\
\midrule
\multirow{6}{*}{Student}&\multirow{2}{*}{XresNet1d101}&&65.8&89.8&83.8&87.0&85.2&63.1&80.0&65.9&32.3&72.5\\
&&$\surd$&\textbf{68.1}&\textbf{95.2}&\textbf{82.5}&\textbf{91.3}&\textbf{86.8}&\textbf{67.3}&\textbf{83.0}&\textbf{68.8}&\textbf{36.4}&\textbf{75.5}\\
&\multirow{2}{*}{ResNet1d34}&&63.9&92.1&84.5&87.5&85.5&33.8&60.0&65.1&26.9&66.6\\
&&$\surd$&\textbf{69.1}&\textbf{93.3}&\textbf{86.3}&\textbf{87.5}&\textbf{87.3}&\textbf{50.0}&\textbf{74.5}&\textbf{66.3}&\textbf{30.8}&\textbf{71.7}\\
\bottomrule[1.5pt]
\end{tabular}
}
\end{table}

\begin{table}[H]
\centering
\caption{MVKT’s results of different diseases on PTB-XL dataset.}
\label{table5}
\vskip 0.1in
\resizebox{0.65\textwidth}{!}{%
\begin{tabular}{ccccccccc}
\toprule
&\multirow{2}[2]{*}{Backbone}&\multirow{2}[2]{*}{MVKT}&\multicolumn{6}{c}{F1\_score}\\
\cmidrule{4-9}
&&&NORM&MI&STTC&CD&HYP&Average\\
\midrule[1.5pt]
Teacher&CNN\_Ag&&74.8&60.5&73.2&86.2&76.0&74.2\\
\midrule
\multirow{2}{*}{Student}&\multirow{2}{*}{CNN\_Ag}&&58.9&45.8&56.3&79.7&66.5&61.4\\
&&$\surd$&\textbf{60.1}&\textbf{46.4}&\textbf{57.4}&\textbf{80.3}&\textbf{67.8}&\textbf{62.4}\\
\midrule[1.5pt]
Teacher&ResNet1d34&&74.5&58.6&75.1&85.8&76.8&74.2\\
\midrule
\multirow{2}{*}{Student}&\multirow{2}{*}{ResNet1d34}&&59.2&45.0&55.4&80.1&66.4&61.2\\
&&$\surd$&\textbf{61.3}&\textbf{47.3}&\textbf{56.7}&\textbf{80.6}&\textbf{68.7}&\textbf{62.9}\\
\midrule[1.5pt]
Teacher&ResNet1d\_wang&&74.9&58.7&75.9&86.6&76.8&74.6\\
\midrule
\multirow{2}{*}{Student}&\multirow{2}{*}{ResNet1d\_wang}&&60.2&46.9&56.0&80.8&67.2&62.2\\
&&$\surd$&\textbf{60.4}&{46.4}&\textbf{57.7}&\textbf{80.9}&\textbf{68.5}&\textbf{62.5}\\
\midrule[1.5pt]
Teacher&Inception1d&&76.6&60.1&76.1&86.4&76.0&75.1\\
\midrule
\multirow{2}{*}{Student}&\multirow{2}{*}{Inception1d}&&58.8&46.3&56.2&80.3&67.0&61.7\\
&&$\surd$&\textbf{60.2}&\textbf{47.8}&\textbf{57.4}&\textbf{80.6}&\textbf{66.5}&\textbf{62.5}\\
\midrule[1.5pt]
Teacher&XresNet1d101&&74.8&53.8&72.7&85.9&75.6&72.6\\
\midrule
\multirow{2}{*}{Student}&\multirow{2}{*}{XresNet1d101}&&58.2&45.5&55.6&80.5&66.6&61.3\\
&&$\surd$&\textbf{59.8}&\textbf{47.2}&\textbf{57.3}&\textbf{80.4}&\textbf{67.5}&\textbf{62.4}\\
\bottomrule[1.5pt]
\end{tabular}
}
\end{table}
\section{Illustration of Multi-label diseases Knowledge Distillation}
\begin{figure}[h]
\centering
\includegraphics[width=0.95\columnwidth]{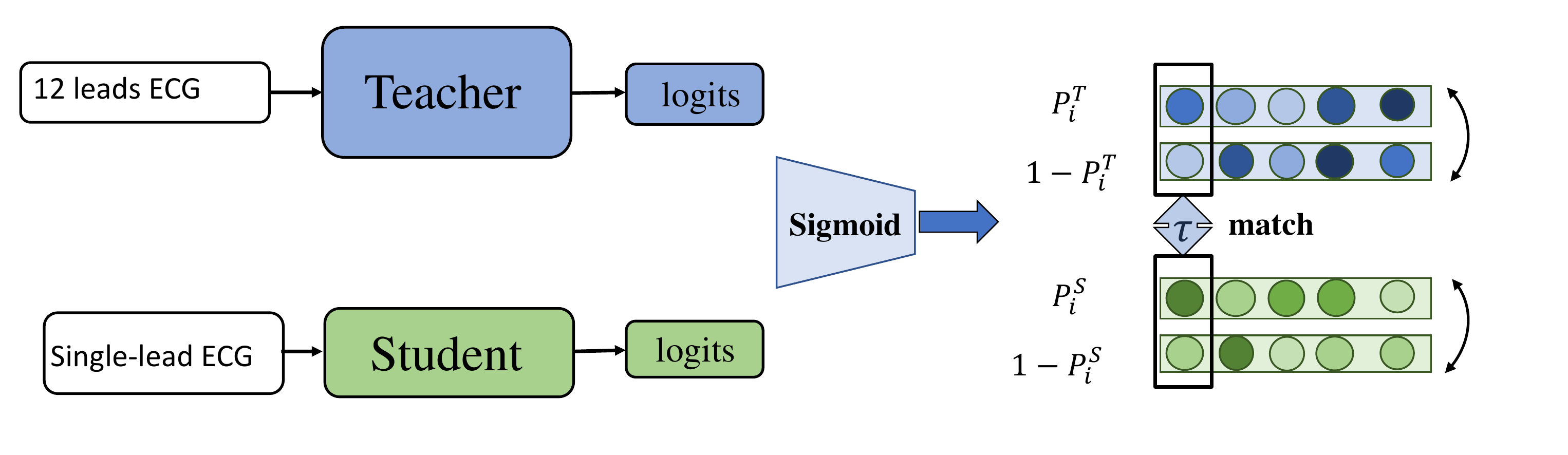} 
\caption{{llustration of Multi-label diseases Knowledge Distillation.}}
\label{fig5}
\end{figure}

\end{document}